\documentclass[11pt]{article}
\usepackage[preprint]{acl}
\usepackage{times,latexsym}
\usepackage[T1]{fontenc}
\usepackage[utf8]{inputenc}
\usepackage{newunicodechar}
\newunicodechar{“}{``}
\newunicodechar{”}{''}
\usepackage{microtype,inconsolata}
\usepackage{graphicx,booktabs,amsmath,amssymb,multirow}
\graphicspath{{figures/}}
\newcommand{\none}{\textsc{None}}
\newcommand{\ind}{\mathbf{1}}
\title{When the Right Answer Is Missing:\\An Arithmetic-Dependent Rejection Bottleneck in Jev}
\author{
Jike Zhong\textsuperscript{1}\thanks{Authors contributed equally. Correspondence to \texttt{jikezhon@usc.edu}.}\quad
Ming Li\textsuperscript{2}\footnotemark[1]\quad
Yuxiang Lai\textsuperscript{3}\footnotemark[1]\\
\normalfont\textsuperscript{1}University of Southern California\\
\normalfont\textsuperscript{2}University of Florida\\
\normalfont\textsuperscript{3}Emory University
}
\hypersetup{pdfauthor={Jike Zhong, Ming Li, Yuxiang Lai},pdftitle={When the Right Answer Is Missing: An Arithmetic-Dependent Rejection Bottleneck in Jev}}
\begin{document}
\maketitle
\begin{abstract}
Typed decision models such as Jev offer an efficient alternative to generative LLMs in decision-making workflows by selecting directly from predefined options. When candidate sets contain no valid answer, TypeSafe recommends including an \emph{"other"} or \emph{“none-of-the-above”} option to enable rejection. In this report, however, we identify an \textbf{arithmetic-dependent rejection bottleneck}: Jev reliably selects correct numerical answers when available but frequently accepts incorrect alternatives when they are absent despite an explicit rejection option. On paired arithmetic problems, answer-present accuracy reaches 99\%, while correct rejection falls to 7\%. Moreover, this gap persists across numerical magnitudes, operation depths, contextual formulations, and rejection labels, and extends to scenarios such as time calculation and capacity rounding.  Yet native Boolean verification achieves 99\% exact-match accuracy on the same answer-absent arithmetic cases, showing that categorical rejection can fail even when the model successfully verifies candidate correctness. Finally, we show that a simple decision threshold selected on separate development problems raises arithmetic rejection accuracy from 7\% to 79\% while retaining 97\% answer-present accuracy, substantially mitigating the failure without retraining or additional inference.
\end{abstract}

\section{Introduction}
Typed decision models such as Jev \cite{typesafe2026} have emerged as an efficient alternative to generative large language models (LLMs) \citep{openai2023gpt4,geminiteam2023gemini,anthropic2024claude} for decision-making workflows. Unlike generative LLMs that produce token sequences autoregressively, Jev accepts a description of the current state and predefined output criteria, then directly returns structured categorical, Boolean, or ordinal judgments \citep{typesafechoice,typesafe2026}. This makes it a natural component for workflows that require bounded decisions rather than generated explanations, such as in record, action selection and policy evaluation.

\begin{figure}[t]
\centering\includegraphics[width=\columnwidth]{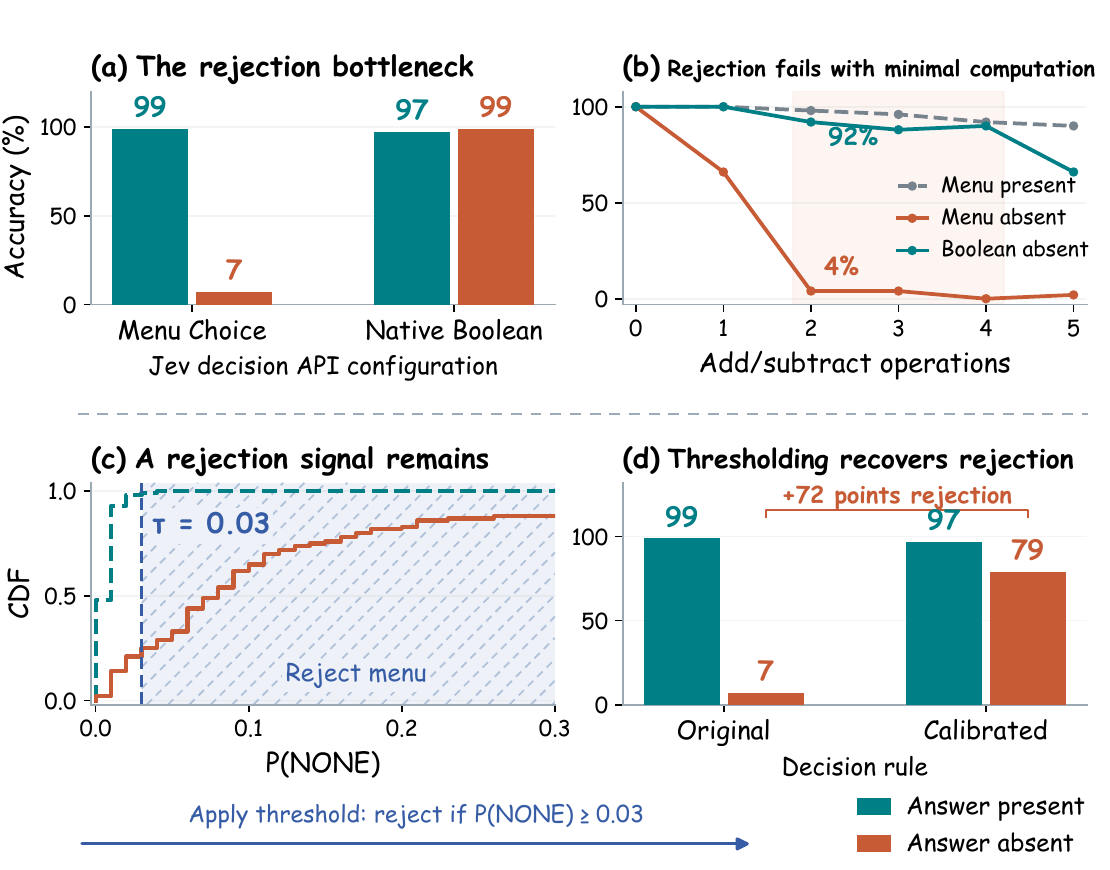}
\caption{\textbf{Arithmetic-dependent rejection and threshold calibration.}
(a) Menu Choice achieves 99\% answer-present accuracy but only 7\% correct rejection even on simple three-operand arithmetic ($a+b-c$), while Boolean verification reaches 99\% answer-absent exact match.
(b) Rejection collapses after just two operations.
(c) \textsc{None} scores retain a useful rejection signal.
(d) A development-selected threshold calibration raises held-out rejection to 79\% while retaining 97\% answer-present accuracy, without additional inference.}
\label{fig:teaser}
\vspace{-10pt}
\end{figure}

However, predefined candidate sets need not contain a valid answer. For example, in payment reconciliation, a system matching a \$100 payment to retrieved invoices must reject all candidates if none has charges minus credits totaling \$100. Recognizing when to reject is therefore essential for reliable decisions, as emphasized by research on LLM abstention \citep{wen2025limits,madhusudhan2025abstention,tomani2024uncertainty,kirichenko2025abstentionbench,wang2025least}. To support rejection, TypeSafe recommends including an \emph{other} or \emph{none-of-the-above} option \citep{typesafechoice}. This raises a basic question: \emph{if a model can select the correct answer, can it also recognize when every supplied answer is wrong?}

We study this question through paired answer-present and answer-absent menus with exact ground truth. Surprisingly, Jev achieves \textbf{99\%} answer-present accuracy on bare arithmetic, even simple two-operation arithmetic ($a+b-c$), yet correctly rejects only \textbf{7\%} of answer-absent menus despite an explicit \emph{``None"} option (\autoref{fig:teaser}a). This rejection gap extends beyond bare expressions to arithmetic-dependent scenarios, including inventory updates, purchase totals, and time calculations. Moreover, rejection already collapses with just \emph{two} elementary operations, showing that the failure arises even under \textbf{minimal computational} demands (\autoref{fig:teaser}b). Together, these results highlight why high \emph{selection accuracy alone} is an incomplete measure of reliability.

To identify the source of this bottleneck, we examine both the required computation and the decision interface. Native Boolean verification correctly rejects all three candidates on 99\% of answer-absent arithmetic cases, demonstrating that Jev \textbf{can} evaluate their numerical correctness. Similarly, providing the correct numerical result directly restores perfect Menu Choice rejection with the same candidates. We further introduce candidate-level T/F Choice, which preserves the Boolean verification questions but uses categorical outputs. Its substantially lower accuracy shows that individual verification alone does not resolve the failure; the \textbf{decision interface} remains consequential.

We therefore ask whether rejection can be improved within Menu Choice itself and find that its output scores retain a useful rejection signal even when the selected answer is incorrect (\autoref{fig:teaser}c). Motivated by this, we propose a simple threshold calibration and show that a threshold selected on separate development problems raises arithmetic rejection from 7\% to 79\% while retaining 97\% answer-present accuracy, without retraining or additional inference (\autoref{fig:teaser}d).

\paragraph{Contributions.}
\begin{itemize}
    \item We identify an \textbf{arithmetic-dependent rejection bottleneck}: Jev reliably selects valid answers but frequently fails to reject invalid candidate sets.
    \item We conduct extensive ablations, controlled comparisons, and error analyses to characterize the roles of computation, decision interface, and task formulation.
    \item We propose a simple \textbf{threshold calibration} that substantially improves rejection while retaining high selection accuracy, without retraining or additional inference.
\end{itemize}

\begin{figure}[t]
\centering\includegraphics[width=\columnwidth]{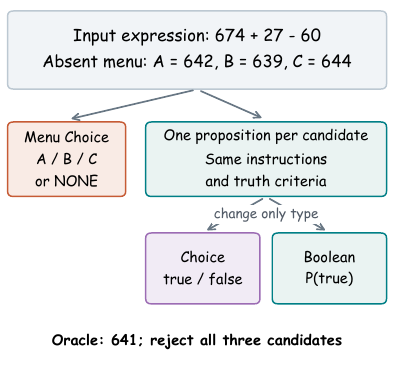}
\caption{\textbf{Three formulations of the same decision.} Menu Choice selects a candidate or \none{}. Our T/F Choice ablation verifies each candidate through the categorical API; native Boolean changes only the question type. Candidate questions share one request. The oracle answer is shown for illustration and is not supplied in the arithmetic input.}
\label{fig:protocol}
\vspace{-10pt}
\end{figure}

\section{Related Work}
\paragraph{Selection and rejection.}
Prior work examines least-incorrect answer selection \citep{wang2025least} and domain-dependent degradation under none-of-the-above questions \citep{tam2025none}. More broadly, studies of LLM abstention document failures on unanswerable questions \citep{madhusudhan2025abstention,kirichenko2025abstentionbench,wen2025limits} and explore uncertainty-based rejection \citep{tomani2024uncertainty}. We study this distinction in typed decision APIs, isolating an arithmetic-dependent rejection bottleneck through matched interface comparisons and computation controls, then evaluating threshold calibration as a repair.

\begin{figure*}[t]
\centering\includegraphics[width=\textwidth]{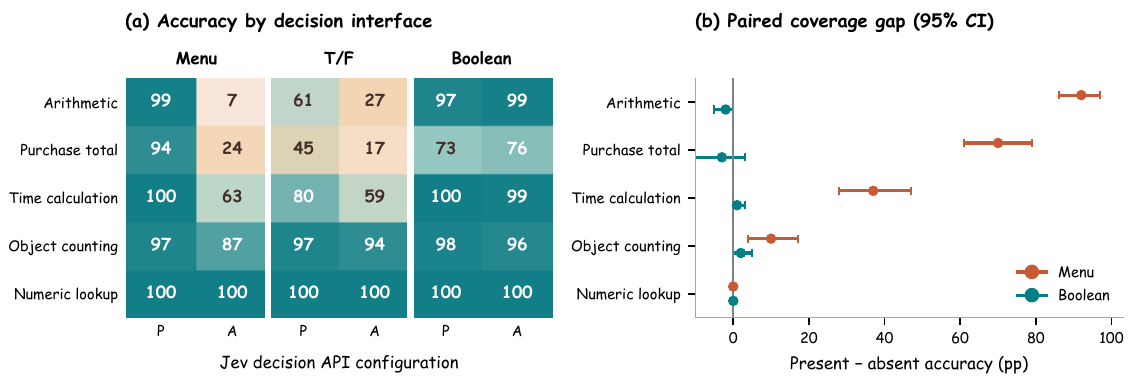}
\caption{\textbf{Strong selection conceals task-dependent rejection failures.}
(a) Accuracy on 100 paired problems per task; P/A denotes answer present/absent. T/F and Boolean require all three judgments correct. Time and counting pool 20 screening and 80 fresh cases.
(b) Present-minus-absent gaps with 95\% paired-bootstrap intervals. Menu rejection deteriorates most on arithmetic and purchase totals; lookup eliminates the gap.}
\label{fig:main}
\vspace{-10pt}
\end{figure*}

\paragraph{Jev interface behavior.}
TypeSafe documents mathematical limitations and Boolean/Choice inconsistencies \citep{typesafe2026}. \citet{sun2026typesafe} study option-name/rubric conflicts, while \citet{zhang2026scores} examine request configuration on ContractNLI. In contrast, we isolate exact numerical rejection under explicit false and \none{} alternatives. Alternative-label controls connect this analysis to schema sensitivity without reducing the phenomenon to a particular identifier.

\section{Evaluation Framework}
\label{sec:framework}
\subsection{Exact Decisions with Incomplete Menus}
Let $x_j$ denote the facts and rule for case $j$, and let $y_j=f(x_j)$ be the deterministically computed answer. Each case has a three-candidate menu $C_{jc}=(a_1,a_2,a_3)$ with coverage $c\in\{0,1\}$. Candidate correctness is
\begin{equation}
 z_{jci}=\ind[a_i=y_j],\qquad \sum_{i=1}^{3}z_{jci}=c.
\end{equation}
An answer-present menu has exactly one correct candidate; an answer-absent menu has none. The model must select the correct candidate or reject the menu. Rejection means that the supplied candidates are incorrect, not that the underlying problem is unanswerable.

\subsection{Three Decision Formulations}
\autoref{fig:protocol} illustrates our three formulations on one saved arithmetic case. They use two native API types, Choice and Boolean, with a third Candidate T/F Choice as an ablation constructed using the Choice type, rather than a third native primitive.

\paragraph{Menu Choice.}
A categorical question selects from $\{A,B,C,\none{}\}$, where each option has an explicit correctness criterion. Following TypeSafe's recommendation, \none{} denotes that no valid candidate \citep{typesafechoice}. We score the API-returned choice $\widehat y$ against the correct option $y$.

\paragraph{Candidate T/F Choice.}
To test whether \textbf{individual verification} repairs rejection, our diagnostic ablation asks whether each candidate is correct using categorical \texttt{true}/\texttt{false} options, batching all three questions in one request. We predict $\widehat z_i^C=\ind[q_i\geq 0.5]$, where $q_i$ is candidate $i$'s \texttt{true} probability. A question is considered correct only if all three labels are correct.

\paragraph{Native Boolean.}
To isolate the effect of \textbf{decision type}, we submit the same candidate questions with Boolean type and predict $\widehat z_i^B=\ind[b_i\geq 0.5]$, where $b_i$ is the returned truth probability. Matched payloads differ only in the \texttt{type} field, verified programmatically. Similarly, all three labels must be correct.

\begin{figure*}[t]
\centering\includegraphics[width=\textwidth]{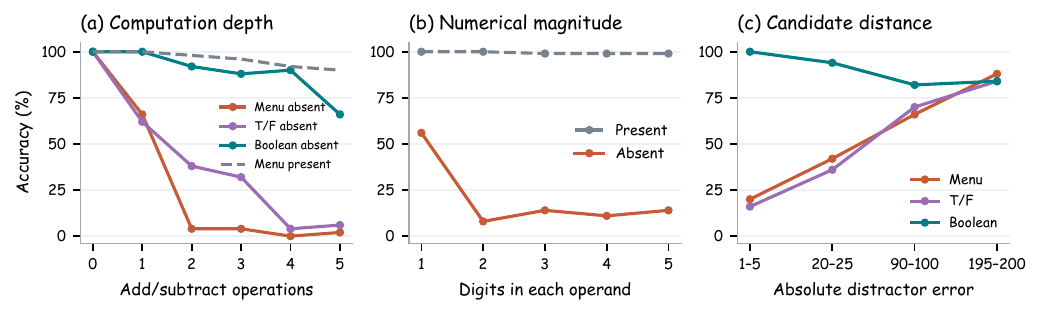}
\caption{\textbf{Rejection fails with simple arithmetic and depends on candidate distance.}
(a) Fifty paired cases per operation depth; depth zero supplies the result.
(b) One hundred fresh two-operation cases per magnitude band.
(c) Fifty paired inventory cases: larger distractor errors improve categorical rejection but not Boolean verification.
Depth and distance vary within cases; magnitude uses independent cases.}
\label{fig:sensitivity}
\vspace{-10pt}
\end{figure*}

\subsection{Metrics and Statistical Unit}
\paragraph{Case-level correctness.}
For candidate verification, let $S=\{i:\widehat z_i=1\}$. A single positive selects that candidate; no positives rejects the menu; multiple positives count as incorrect without score-based tie-breaking. Since each menu has at most one valid candidate, decision accuracy equals \textbf{all-candidate exact match}: $A_{h,c}=\frac{1}{n}\sum_{j=1}^{n}\ind[\widehat{\boldsymbol z}_{jc}^{h}=\boldsymbol z_{jc}]$ for interface $h$ and coverage condition $c$.

\paragraph{Paired comparisons.}
The coverage gap $G_h=A_{h,1}-A_{h,0}$ compares answer-present and answer-absent accuracy on paired problems. Interface and distance comparisons likewise preserve pairing. Named contrasts use 20,000 case-bootstrap resamples (seed 2026092901). AUC describes candidate-score discrimination; accuracy and uncertainty use the underlying problem as the statistical unit. Pilot results report explicit denominators.

\section{Experiments and Results}
\label{sec:experiments}
\subsection{Experimental Setup}
\paragraph{Tasks and paired menus.}
Our main evaluation uses 100 problems each for two-operation arithmetic ($a+b-c$), numerical lookup, and purchase totals ($qp+s$). Arithmetic samples $a\in[300,999]$ and $b,c\in[10,99]$; purchase totals use $q\in[3,11]$, $p\in[11,59]$, and $s\in[5,24]$. Present menus use answer offsets $\{-2,0,+3\}$ and absent menus use $\{-2,+1,+3\}$, preserving the facts and two distractors while shuffling candidate order within each coverage condition. All interfaces receive the same menus; arithmetic and lookup additionally share answers and candidate text. Contextual controls comprise 128 inventory problems and 64 problems each for refund policy and logical eligibility. Subsequent experiments vary computation depth, numerical magnitude, candidate distance, operator structure, task family, and rejection encoding.

\paragraph{Models and scoring.}
We evaluate Jev and record structured predictions and probabilities. Programmatic generators provide exact ground truth. Menu accuracy scores the selected option. Candidate verification requires all three truth labels to be correct. We additionally use Qwen3.5-9B (w/ Reasoning) as a generative LLM control on the same menus.

\subsection{Main Results: The Rejection Bottleneck}
Menu Choice achieves 99\% answer-present accuracy on simple arithmetic but only 7\% correct rejection, a 92-point paired gap. Purchase totals show the same pattern: 94\% versus 24\%, a 70-point gap (\autoref{fig:main}). Thus, successful selection does not imply reliable rejection when the correct answer is missing.

Contextual inventory extends the effect beyond bare expressions, with 127/128 correct selections but only 14/128 correct rejections. In contrast, refund-policy and logical-eligibility controls achieve 64/64 under both coverage conditions across all tested interfaces. The bottleneck therefore depends on the task rather than arising whenever a rejection option is needed. These earlier contextual categorical-verification controls use neutral identifiers rather than literal T/F labels.

\subsection{Isolating Computation and Interface Effects}
\paragraph{Removing computation restores rejection.}
To isolate computation from numerical candidate matching, we supply the correct result directly while preserving the arithmetic answers and menus. All three interfaces achieve 100\% accuracy under both coverage conditions. For Menu Choice, rejection rises from 7\% to 100\%: the same numerical candidates support reliable rejection when calculation is removed.

\paragraph{Individual verification does not close the interface gap.}
To separate decomposition from question type, we compare candidate-level T/F Choice with native Boolean, changing only the \texttt{type} field in matched payloads. On answer-absent arithmetic, exact-match accuracy rises from 27\% to 99\%; purchase totals show the same direction of improvement. Thus checking candidates individually does not resolve the failure: the same verification questions produce substantially different accuracy across interfaces.
For example, given $674+27-60=641$ and candidates $(642,639,644)$, Menu Choice and T/F Choice both accept 639, whereas Boolean correctly rejects all three. The failure therefore persists under categorical verification even when native Boolean verification succeeds.

\begin{figure}[t]
\centering\includegraphics[width=\columnwidth]{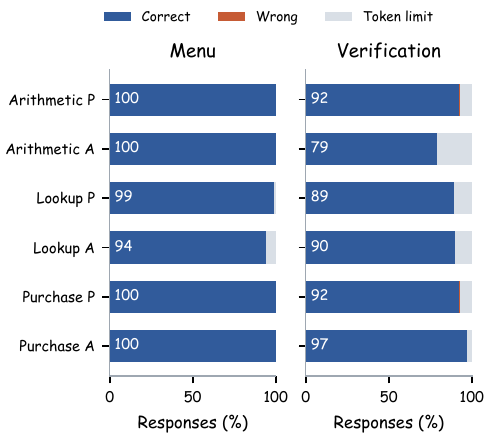}
\caption{\textbf{A capable generative control.} Qwen3.5-9B (w/ reasoning) on the same cases per family and coverage. P/A denotes present/absent. Both computational menus are solved perfectly. Token-limit failures account for 66 of 67 errors across menu and joint-verification responses, distinguishing output completion from valid incorrect decisions.}
\label{fig:qwen}
\vspace{-10pt}
\end{figure}

\paragraph{A capable generative model rejects the same menus.}
To test whether the candidate construction itself prevents rejection, we evaluate reasoning-enabled Qwen3.5-9B \citep{qwen2026} using menu and joint-verification prompts on the identical transfer cases. \autoref{fig:qwen} shows that Qwen reaches 100\% present and absent menu accuracy on both arithmetic and purchase totals. Lookup scores 99\%/94\%, with all seven errors caused by truncation. Joint-verification accuracy is 92\%/79\% for arithmetic, 89\%/90\% for lookup, and 92\%/97\% for purchase. Overall, 66 of 67 errors are truncations. The perfect computational-menu results establish that the same candidate sets permit reliable rejection.

\subsection{Depth, Magnitude, and Operator Structure}
\paragraph{Rejection collapses with minimal computation.}
We vary additive depth from zero (the result is supplied) to five operations on 50 paired base cases, preserving answers and candidates. Menu present/absent accuracy falls from 100\%/66\% at one operation to 98\%/4\% at two. Boolean retains 90\%/92\% at two operations and 84\%/90\% at four. At five, Boolean declines to 58\%/66\%, while Menu selection remains 90\% and rejection is 2\%. Categorical rejection thus deteriorates well before Boolean verification (\autoref{fig:sensitivity}a).

\paragraph{Single-digit calculations already expose the gap.}
To isolate magnitude, we generate 100 fresh two-operation $a+b-c$ problems in each band: 1--9, 10--99, 100--999, 1,000--9,999, and 10,000--99,999. Operands, intermediate sums, and answers remain within the band; distractor offsets stay fixed. Selection is 100\%, 100\%, 99\%, 99\%, and 99\%, respectively, whereas rejection is 56\%, 8\%, 14\%, 11\%, and 14\%. Failure therefore occurs even within 1--9 and remains severe, without a monotonic magnitude trend, across larger bands (\autoref{fig:sensitivity}b).

\paragraph{Distance affects interfaces differently.}
To distinguish candidate proximity from problem magnitude, we give 50 shared inventory cases absent candidates at absolute-error bands 1--5, 20--25, 90--100, and 195--200. Menu rejection increases from 20\% to 88\%, and T/F exact rejection from 16\% to 84\%. In contrast, Boolean declines from 100\% to 84\% (\autoref{fig:sensitivity}c). Increasing numerical separation helps categorical rejection but does not produce a common difficulty ordering across interfaces.

\begin{table}[t]
\centering\small
\begin{tabular}{@{}lrr@{}}\toprule
Two-operation pattern & Present & Absent\\\midrule
$a\times b\times c$ &9/10&10/10\\
$a\times b+c$ &9/10&3/10\\
$a\times b-c$ &8/10&1/10\\
$a+b\times c$ &10/10&1/10\\
$a-b\times c$ &8/10&0/10\\\midrule
$a\div b\div c$ &10/10&8/10\\
$a\div b+c$ &6/10&1/10\\
$a\div b-c$ &5/10&2/10\\
$a+b\div c$ &9/10&3/10\\
$a\times b\div c$ &10/10&9/10\\\bottomrule
\end{tabular}
\caption{\textbf{Rejection varies with operator composition.} Multiplication/division chains achieve higher rejection accuracy than expressions combining these operations with addition or subtraction.}
\label{tab:operators}
\vspace{-10pt}
\end{table}

\paragraph{Mixed operators reproduce the rejection gap.}
To test beyond additive arithmetic, we evaluate 50 single multiplications, 50 exact divisions, and 100 two-operation problems (10 per pattern), using standard precedence and exact integer division. Two-operation cases retain the original operand ranges and positive answers. Single multiplication and division retain high rejection accuracy (98\% and 96\%), whereas combining these operators with addition or subtraction reproduces the selection--rejection gap (\autoref{tab:operators}). The bottleneck thus extends beyond additive expressions while remaining sensitive to operator composition.

\subsection{Task Transfer and Rejection Encoding}
\begin{figure}[t]
\centering\includegraphics[width=\columnwidth]{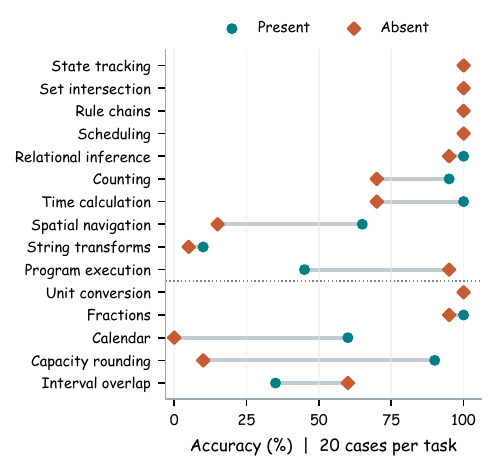}
\caption{\textbf{Rejection gaps extend to other arithmetic-dependent tasks.} Menu Choice results across 15 pilot families, with 20 paired cases each. Time calculation, capacity rounding, and calendar arithmetic show selection--rejection gaps, while several symbolic tasks succeed under both conditions. Time and counting show their initial pilot results; the dotted divider separates the five additional numerical tasks.}
\label{fig:breadth}
\vspace{-10pt}
\end{figure}

\paragraph{The gap extends to other arithmetic-dependent scenarios.}
Beyond bare expressions, we observe selection--rejection gaps in contextual inventory, purchase totals, time calculations, and capacity rounding (\autoref{fig:breadth}). Time calculations reproduce the gap on fresh cases despite perfect answer-present accuracy, while Boolean verification or directly supplying the result largely restores rejection. Capacity rounding and calendar arithmetic also show rejection failures in the smaller pilot sets, whereas object counting exhibits a milder gap. We hypothesize that the underlying arithmetic in these scenarios contributes to the bottleneck, with its severity depending on the task and operator structure.

\paragraph{Broader controls delimit the affected tasks.}
Five additional numerical families receive 20 paired cases each; \autoref{fig:breadth} reports all 15 screens. State tracking, set intersection, rule chains, and scheduling succeed perfectly. Unit conversion scores 20/20 in both conditions, and fraction comparison scores 20/20 versus 19/20. In contrast, capacity rounding scores 18/20 versus 2/20; computing $\lceil96/14\rceil=7$ is a representative example. Calendar arithmetic scores 12/20 versus 0/20. Spatial navigation and string transformation have weaker answer-present performance, while program execution and interval overlap perform better with the answer absent.

\paragraph{Alternative rejection labels preserve the failure.}
To isolate wording, we rerun the same 100 arithmetic cases with \texttt{Other} and \texttt{None of the above}, preserving rejection criteria, candidate order, and inputs. Their present/absent accuracies are 99\%/9\% and 99\%/6\%, compared with the saved \texttt{NONE} baseline of 99\%/7\%. Separately, moving \none{} across four positions recovers only 5, 0, 2, and 0 of 40 selected inventory failures. These results highlight that neither label substitution nor position changes provide a general repair, motivating the score-based calibration studied next.

\section{Error Analysis and Threshold Calibration}
\subsection{Useful Scores, Incorrect Acceptance}
\begin{figure*}[t]
\centering\includegraphics[width=\textwidth]{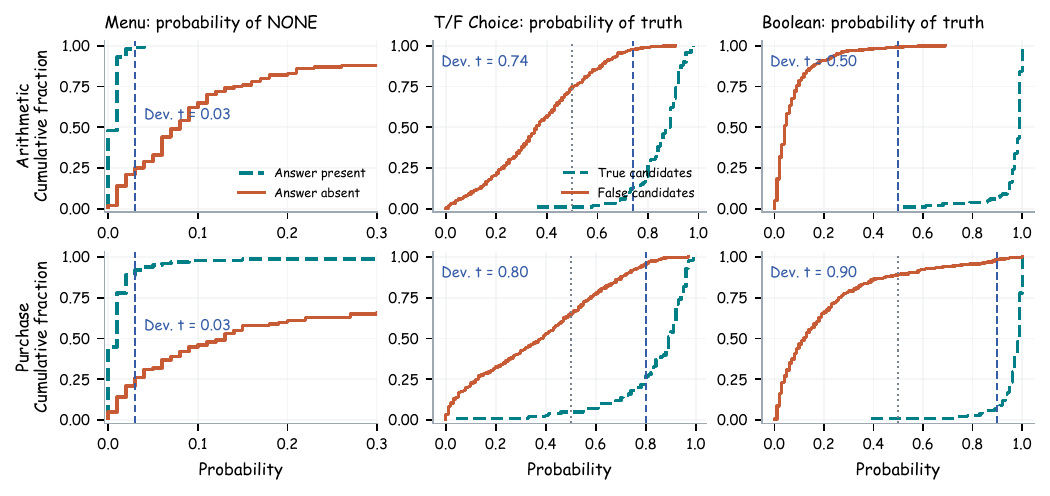}
\caption{\textbf{Score separation motivates threshold calibration.} Empirical CDFs of Menu Choice rejection probabilities (left) and candidate truth scores for T/F Choice and Boolean (center/right). Answer-absent menus tend to receive higher rejection scores, while correct candidates generally receive higher truth scores than incorrect candidates. This separation motivates adjusting decision thresholds to reduce false acceptance. Gray dotted lines mark the default verification threshold; blue dashed lines mark development-selected thresholds.}
\label{fig:errors}
\vspace{-10pt}
\end{figure*}
The preceding controls show that the decision interface affects rejection. We next ask whether the returned scores retain enough information to repair these errors without changing the interface. In \autoref{fig:errors}, arithmetic T/F Choice often assigns incorrect candidates truth scores above .5, although correct candidates generally receive higher scores. We quantify this separation using the area under the receiver operating characteristic curve (AUC): the probability that a randomly chosen correct candidate scores above a randomly chosen incorrect candidate, with half credit for ties. An AUC of .5 indicates chance-level ordering and 1 indicates perfect ordering. Arithmetic T/F achieves an AUC of .985 despite only 88/200 exact-match menus; purchase T/F shows the same separation between ranking and decisions, with an AUC of .942 and 62/200 exact matches.

This strong ordering suggests that threshold placement contributes to the errors. Under the default .5 threshold, arithmetic T/F accepts multiple candidates on 55/200 menus, compared with 3/200 for Boolean; purchase exhibits the same pattern. Selecting the highest-scoring candidate would hide these conflicting judgments and would still require a separate rejection rule when every candidate is wrong. Instead, the score separation motivates calibrating the acceptance threshold while retaining all-candidate exact-match evaluation.

Menu Choice offers a related opportunity through a different score. Answer-absent menus often assign more probability to \none{} than answer-present menus, even when \none{} does not outrank the numerical alternatives. This motivates thresholding rejection probability directly. Thus candidate verification and menu selection call for different calibration rules, both using scores already returned by the model.

\subsection{A Development-Selected Decision Rule}
\paragraph{Menu rejection threshold.}
Let $p_\varnothing$ be the probability of \none{} and $p_i$ the candidate probabilities. We replace categorical argmax with
\begin{equation}
\widehat d_\tau(x,C)=
\begin{cases}
\varnothing,&p_\varnothing\geq\tau,\\
\arg\max_{i\in\{1,2,3\}}p_i,&p_\varnothing<\tau.
\end{cases}
\label{eq:menu}
\end{equation}
This rule checks whether the menu should be rejected before selecting a candidate, allowing rejection without requiring \none{} to have the largest probability.

\paragraph{Candidate acceptance threshold.}
For candidate verification, we instead threshold each T/F or Boolean truth score $s_i$: $\widehat z_i=\ind[s_i\geq\tau]$ and retain all-three exact-match scoring. A higher threshold targets the false acceptances identified above; one positive selects a candidate, no positives rejects the menu, and multiple positives remain an error.

\paragraph{Calibration protocol.}
To select an operating point that balances selection and rejection, we search $\tau\in\{0,.01,\ldots,1\}$ separately for each task/interface on 20 development problems, each with present and absent menus. The objective is mean case accuracy across the two balanced conditions; ties favor proximity to .5, then the smaller threshold. Arithmetic, lookup, and purchase use the earlier 20-case pilot for development and the disjoint 100-case transfer set for evaluation. Time and counting use their initial 20 for development and fresh 80 for evaluation. All calibration is performed offline on saved scores, and each selected threshold is applied unchanged to its evaluation set, requiring neither retraining nor additional inference.

\begin{figure*}[t]
\centering\includegraphics[width=\textwidth]{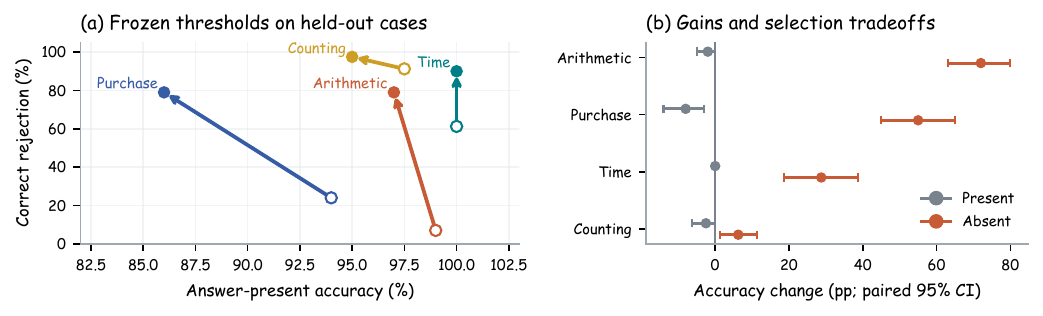}
\caption{\textbf{Threshold adjustment recovers rejection on separate cases.} (a) Arrows connect original menu decisions (open circles) to calibrated decisions (filled circles). Upward movement improves rejection; leftward movement loses selection accuracy. (b) Paired 95\% bootstrap intervals quantify both effects. Arithmetic/purchase use 100 test cases; time/counting use 80. Thresholds are chosen only on the corresponding 20-case development sets.}
\label{fig:calibration}
\end{figure*}

\subsection{Held-Out Recovery and Tradeoffs}
\begin{table}[t]
\centering\small
\begin{tabular}{@{}llrrr@{}}\toprule
Task & Interface & $\tau$ & Present & Absent\\\midrule
Arithmetic & Menu &.03&99$\to$97&7$\to$79\\
 & T/F &.74&61$\to$86&27$\to$90\\
 & Boolean &.50&97$\to$97&99$\to$99\\
Purchase & Menu &.03&94$\to$86&24$\to$79\\
 & T/F &.80&45$\to$73&17$\to$82\\
 & Boolean &.90&73$\to$87&76$\to$97\\\midrule
Time & Menu &.13&80$\to$80&49$\to$72\\
 & T/F &.66&61$\to$59&49$\to$65\\
 & Boolean &.50&80$\to$80&79$\to$79\\
Counting & Menu &.09&78$\to$76&73$\to$78\\
 & T/F &.79&78$\to$78&76$\to$77\\
 & Boolean &.50&79$\to$79&77$\to$77\\\bottomrule
\end{tabular}
\caption{\textbf{Original $\to$ calibrated correct counts.} Arithmetic and purchase denominators are 100 per coverage; time and counting denominators are 80. Lookup remains 100/100 throughout at $\tau=.5$. No evaluation labels are used to select thresholds.}
\label{tab:cal}
\vspace{-10pt}
\end{table}
\autoref{tab:cal} and \autoref{fig:calibration} show that these development-selected thresholds recover rejection on held-out cases. For arithmetic menus, $\tau=.03$ raises correct rejection from 7\% to 79\% while retaining 97\% answer-present accuracy, compared with 99\% originally. Balanced accuracy therefore rises from 53\% to 88\%: the rejection gains substantially exceed the selection losses.

Candidate-level calibration likewise converts the strong score ordering into more accurate decisions. With $\tau=.74$, arithmetic T/F exact match rises from 61\%/27\% to 86\%/90\% on present/absent menus. Boolean retains its default threshold of .5 and its 97\%/99\% accuracy. These results connect the score separation in \autoref{fig:errors} to a practical correction of false acceptance.

The same intervention also improves rejection beyond bare arithmetic. On fresh time cases, menu calibration increases rejection from 49/80 to 72/80 while preserving perfect selection. Purchase balanced accuracy improves from 59\% to 82.5\%, exchanging eight present-case successes for 55 absent-case rescues; counting exchanges two for five. Threshold calibration thus offers substantial recovery where the scores separate valid from invalid menus, with selection--rejection tradeoffs that vary by task.

\section{Discussion and Conclusion}
\paragraph{Computation and rejection interact.}
Strong selection and Boolean verification show that Jev can solve the underlying problems, while supplying computed answers restores Menu Choice rejection. However, matched T/F Choice does not recover Boolean accuracy, showing that decomposition alone is insufficient. Together, these controls link the bottleneck to computation and decision interface. The gap extends to contextual arithmetic, time calculation, and capacity rounding, while successful lookup, symbolic controls, and multiplication/division chains reveal its task dependence.

\paragraph{Implications for decision workflows.}
In payment reconciliation and similar workflows, retrieved candidates may all be invalid. An explicit rejection option therefore needs evaluation on paired answer-present and answer-absent cases. Boolean verification changes the interface, whereas threshold calibration improves decisions from existing scores without retraining or additional inference. Their effectiveness should be assessed through both rejection gains and selection losses; deterministic computation offers another remedy when structured inputs permit it.

\paragraph{Conclusion.}
In this report, we identify an arithmetic-dependent rejection bottleneck in Jev despite strong selection accuracy. Through paired coverage tests, matched interfaces, and lookup controls, we show that categorical rejection can fail even when candidate verification succeeds. Motivated by score-distribution analysis, our threshold calibration substantially improves held-out rejection without retraining or additional inference. These findings support evaluating selection and rejection jointly. Future work includes real-world retrieval settings, cross-task calibration, and training for reliable rejection.

\clearpage
\section*{Limitations}
We evaluate Jev through its hosted API without access to model weights or internal activations. Our experiments and interpretations therefore characterize observable behavior through controlled inputs and returned outputs, rather than directly identifying the internal mechanisms behind the rejection bottleneck.
\section*{Ethics Statement}
All experimental inputs are synthetic numerical or symbolic records without personal data. The study evaluates output reliability and does not execute financial, physical, or account actions. Its intended use is to improve validation of automated decisions and to make rejection errors visible alongside successful selection.

\bibliography{references}
\end{document}